\documentclass[12pt]{article}

\usepackage{amsmath}
\usepackage{times}
\usepackage{graphicx}
\usepackage{color}
\usepackage{multirow}
\usepackage[authoryear]{natbib}
\usepackage{rotating}
\usepackage{bbm}
\usepackage{latexsym}

\newcommand{\captionfonts}{\normalsize}

\makeatletter  
\long\def\@makecaption#1#2{%
  \vskip\abovecaptionskip
  \sbox\@tempboxa{{\captionfonts #1: #2}}%
  \ifdim \wd\@tempboxa >\hsize
    {\captionfonts #1: #2\par}
  \else
    \hbox to\hsize{\hfil\box\@tempboxa\hfil}%
  \fi
  \vskip\belowcaptionskip}
\makeatother

\def\eTP{{\em tp}}
\def\eFP{{\em fp}}
\def\eTN{{\em tn}}
\def\eFN{{\em fn}}

\def\bd{{\mbox{\boldmath $d$}}}
\def\bmu{{\mbox{\boldmath $\mu$}}}
\def\bw{{\mbox{\boldmath $w$}}}
\def\bx{{\mbox{\boldmath $x$}}}
\def\bzero{{\mbox{\boldmath $0$}}}

\def\mE{{\mbox{\bf E}}}
\def\mH{{\mbox{\bf H}}}
\def\mS{{\mbox{\bf S}}}

\def\tct {{\sf c4.5}}
\def\tsvmone {{\sf svm1}}
\def\tsvmtwo {{\sf svm2}}
\def\tlda {{\sf lda}}
\def\tqda {{\sf qda}}
\def\tknn {{\sf knn}}
\def\trf {{\sf rf}}

\begin{document}

\hspace{13.9cm}1

\ \vspace{20mm}\\

\noindent {\LARGE Multivariate Comparison of Classification Algorithms\footnote{Preliminary version of this work appeared as \citep{lion5}.}}

\ \\
{\large Olcay~Taner~Y{\i}ld{\i}z$^{\displaystyle 1}$, Ethem~Alpayd{\i}n$^{\displaystyle 2}$}\\
{$^{\displaystyle 1}$Department of Computer Engineering, I{\c s}{\i}k University, TR-34980, {\.I}stanbul Turkey}\\
{$^{\displaystyle 2}$Department of Computer Engineering, Bo{\u g}azi{\c c}i University, TR-34342, {\.I}stanbul, Turkey}\\
%

%\ \\[-2mm]
\noindent {\bf Keywords:} Statistical Tests, Design of Experiments, Multivariate Analysis.

\thispagestyle{empty}
\ \vspace{-0mm}\\
%
%Abstract
\begin{center} {\bf Abstract} \end{center}
Statistical tests that compare classification algorithms are univariate and use a single performance measure, e.g., misclassification error, $F$ measure, AUC, and so on. In multivariate tests, comparison is done using multiple measures simultaneously. For example, error is the sum of false positives and false negatives and a univariate test on error cannot make a distinction between these two sources, but a 2-variate test can. Similarly, instead of combining precision and recall in $F$ measure, we can have a 2-variate test on (precision, recall). We use Hotelling's multivariate $T^2$ test for comparing two algorithms, and when we have three or more algorithms we use the multivariate analysis of variance (MANOVA) followed by pairwise post hoc tests. In our experiments, we see that multivariate tests have higher power than univariate tests, that is, they can detect differences that univariate tests cannot. We also discuss how multivariate analysis allows us to automatically extract performance measures that best distinguish the behavior of multiple algorithms.
%%%%%%%%%%%

\section{Introduction}

In many applications, we have several candidate classification algorithms and we need to choose the best one. Many performance measures have been proposed in the literature, each with its domain of application. For example, in classification problems, misclassification error is used. In information retrieval, the measures used are precision and recall, and in signal detection, the measures are true positive rate (tpr) and false positive rate (fpr).

Misclassification error is the sum of false positives and false negatives, and because it is a simple sum, it does not make a distinction between these two sources of error. Similarly, $F$ measure combines precision and recall in a single number. Using a single measure is good because it is simple; it can be plotted, visually analyzed, and a univariate test can be defined on it. However, it leads to a loss of information. For example, a comparison using misclassification error cannot make a distinction between its two sources of false positives and false negatives. Two classifiers may have the same error, but one may have all its error due to false positives, the other due to false negatives, and we will not be able to detect this difference if the comparison metric is simply the error.

This difference is critical in many applications. For example, let us say we want to check if a patient has cancer or not. Then a false positive corresponds to saying that a healthy patient has cancer, and a false negative corresponds to discharging a patient with cancer. It is clear that these two are two different types of error and should not be just summed up as if they are interchangeable, but should be considered separately.

One possibility is to assign different loss values to false positives and false negatives, then calculate and compare the expected risks; but expected risk is a weighted sum and two classifiers that have different false positives and negatives may have the same expected risk. Or one can plot a ROC curve, calculate the area under it and compare using those; but again note that two classifiers may have different ROC curves but have the same area under the curve values. Such approaches all correspond to a summing up and hence condensing of multiple results into a single value and hence lose information.

What we discuss here is a {\em multivariate test\/} that can do comparison using multiple measures simultaneously, without needing to combine them in a single value, i.e., error, $F$ measure, expected risk, area under the ROC curve, and so on. That is, we collect $p$-dimensional {\em multivariate\/} ($p\ge 2$) statistics. For example, we can have a 2-dimensional vector of (false positives, false negatives) or (precision, recall), and do a 2-variate test.

We proposed the use of multivariate testing in our previous work \citep{lion5}; here, we extend the discussion of multivariate testing, propose new uses for different type of settings, and also show how multivariate analysis can be used to define new application-specific measures of performance.

This paper is organized as follows: In Section~\ref{sec-tests}, we discuss the multivariate tests for comparing two or more algorithms, generalizing from univariate comparison. Our case studies and experimental results on seven algorithms and five data sets are given in Section~\ref{sec-exp}. We conclude and discuss future work in Section~\ref{sec-conc}.

\def\etp{{\em tp}}
\def\efn{{\em fn}}
\def\efp{{\em fp}}
\def\etn{{\em tn}}

\section{Multivariate Tests}
\label{sec-tests}

In statistical testing, when we compare two populations in a paired manner, we calculate paired differences and test a hypothesis on these differences. When we compare three or more populations, we define a test based on differences of populations from a global average \citep{ozanetal12}.

For example, we may have two algorithms that we want to compare on a data set in terms of some performance metric(s); for example, we may want to compare a decision tree and a $k$-nearest neighbor classifier. This is the most frequently used scenario. Or, we may want to test two variants of the same algorithm; for example, we may want to compare the polynomial kernel and the Gaussian kernel with support vector machines. 

Sometimes, we have $L>2$ algorithms that we want to compare on a single data set in terms of some metrics. These may be different algorithms or different variants of the same algorithm; for example, we may be interested in comparing $L>2$ different kernels for support vector machines.

On a single instance, a classifier makes a decision that is true or false, and the number of errors that the classifier does on a set of $N$ instances is binomially distributed, which is approximately normal due to the central limit theorem, unless $N$ is very small. Measures such as tpr, precision, and so on are all calculated similarly and hence approximate normality also holds for them. Indeed, parametric tests that assume normality are frequently used \citep{dietterich98,alpaydin99}. 

The central limit theorem also holds for the multivariate case and the $p$-dimensional statistics are approximately $p$-dimensional multivariate normal distributed. Their mean is a $p$-dimensional vector that corresponds to the expected performance in terms of the $p$ measures used and the $p\times p$ covariance matrix corresponds to the correlations between these measures.

Below, first we discuss the hypothesis test for comparison of two algorithms, and then the one for comparison of three or more.

\subsection{Comparing Two Algorithms}
\label{sec-pair}

Let us say we have two classification algorithms. We train both on the same $k$ training folds, then test on the same validation data folds and calculate the resulting $k$ separate $2\times 2$ confusion matrices $M_{ij}, i=1,2, j=1,\ldots,k$, containing {\etp}, {\efn}, {\efp}, and {\etn}, namely, true positives, false negatives, false positives, and true negatives. From these confusion matrices, we calculate the $p$ performance measures we are interested in. For example, if we compare in terms of (tpr, fpr) or (precision, recall), then $p=2$  \citep{lion5}.

The hypothesis test checks whether the two samples are drawn from two normal populations with the same mean, or equivalently, whether the sample of paired differences are drawn from a population with zero mean. Let us say $\bx_{ij}\in\Re^p, i=1,2, j=1,\ldots,k$ is the performance vector containing $p$ performance values. For the {\em multivariate paired Hotelling's test}, we calculate {\em paired differences\/} $ \bd_j=\bx_{1j}-\bx_{2j}$ and check if they are drawn from a $p$-variate Gaussian with zero mean:
$$H_0: \bmu_d=\bzero \mbox{\ vs.\ } H_1: \bmu_d\ne \bzero.$$ 

We calculate the sample mean and covariance matrix:
$$\overline{\bd}=\frac{1}{k}\sum_{j=1}^ k \bd_j \mbox{,\ \ } \mS_d=\frac{1}{k-1}\sum_j (\bd_j-\overline{\bd})(\bd_j-\overline{\bd})^T.$$

Under the null hypothesis that the two algorithms have the same expected performance, the test statistic 
\begin{equation}
T'^2=k\overline{\bd}^T\mS_d^{-1}\overline{\bd}
\label{eq-multivar-t2}
\end{equation}

\noindent is {\em Hotelling's $T^2$\/} distributed with $p$ and $k-1$ degrees of freedom \citep{rencher95}. For given $\alpha$, we reject the null hypothesis if $T'^2 > T^2_{\alpha,p,k-1}$\footnote{Hotelling's $T^2_{p,m}$ can be approximated using $F$ distribution via the formula 
\[
\left( \frac{m - p + 1}{mp} \right) T^2_{p, m}\sim F_{p, m - p + 1}
\]}.

For $p=1$, this reduces to the well-known (univariate) paired $t$ test. $\overline{\bd}^T\mS_d^{-1}\overline{\bd}$ of equation (\ref{eq-multivar-t2}) measures the (squared) normalized (Mahalanobis) distance to the origin (hypothesized value) and the test rejects if it is too large for given $\alpha$, $p$, and $k$.

If the multivariate test rejects, we can do $p$ separate post hoc univariate paired $t$ tests in each dimension to check the source of difference. For example, if a multivariate test on (precision, recall) rejects, we can check if the difference is due to a significant difference in precision, recall, or both. 

Note that the multivariate test can find a difference even though none of the univariate differences is significant; that is why the multivariate test is more powerful and should be preferred. The linear combination of variates that causes the maximum difference is
\begin{equation}
\bw= \mS_d^{-1}\overline{\bd} 
\label{eq-w-diff}
\end{equation}

Note that this is the Fisher's linear discriminant direction; we are looking for the direction that maximizes the separation of two normal groups.

\subsection{Comparing $L>2$ Algorithms}
\label{sec-L}

If we have $L>2$ algorithms, we test whether they all lead to classifiers with the same expected performance. Again we train all algorithms on the same $k$ training folds, test them on the validation  folds and get the confusion matrices. From the confusion matrices, we calculate the performance values for each algorithm on all folds and given $L$ populations, we test for the equality of their means \citep{lion5}.
\begin{eqnarray}
H_0 &:& \bmu_1=\bmu_2=\cdots=\bmu_L \mbox{\ vs.\ } \nonumber\\
H_1 &:& \bmu_r\ne\bmu_s\mbox{\ for at least one pair } r,s \nonumber
\end{eqnarray}

Let $\bx_{ij},i=1,\ldots,L, j=1,\ldots,k$ denote the $p$-dimensional performance vector of algorithm $i$ on validation fold $j$. Multivariate ANOVA (MANOVA) calculates the two matrices of between- and within-scatter:
\begin{eqnarray}
\mH &=& k \sum_{i=1}^L (\bx_{i\cdot} - \bx_{\cdot\cdot})(\bx_{i\cdot} - \bx_{\cdot\cdot})^T 	\nonumber\\
\mE &=& \sum_{i=1}^L \sum_{j=1}^k (\bx_{ij}-\bx_{i\cdot})(\bx_{ij}-\bx_{i\cdot})^T \nonumber\end{eqnarray}

Then, under the null hypothesis, the test statistic
\begin{equation}
\Lambda' = \frac{|\mE|}{|\mE+\mH|}
\end{equation}

\noindent is {\em Wilks'\/} $\Lambda$ distributed with $p,L(k-1),L-1$ degrees of freedom \citep{rencher95}. We reject $H_0$ if $\Lambda' \le \Lambda_{\alpha,p,L(k-1),L-1}$\footnote{Wilks' $\Lambda$ can be approximated using the chi-square distribution via the formula 
\[
\left(\frac{p - n + 1}{2} - m\right)\log\Lambda_{p, m, n}\sim\chi^2_{np}
\]}. 

We reject if $\Lambda'$ is small: If the sample mean vectors are equal, we expect $\mH$ to be $\bzero$ and $\Lambda'$ to approach 1; as the sample means become more spread, $\mH$ becomes ``larger'' than $\mE$ and $\Lambda'$ approaches 0. In the univariate case ($p=1$), MANOVA reduces to the well-known ANOVA. 

The significant difference may be due to any of the dimensions, which we can pinpoint using post hoc ANOVA on separate dimensions. The difference may also be due to some linear combination of the variates: The mean vectors occupy a space whose dimensionality is $\min(p,L-1)$; its dimensions are the eigenvectors of $\mE^{-1}\mH$ and the corresponding eigenvalues give their importance.

If MANOVA rejects, we can use the pairwise test of equation (\ref{eq-multivar-t2}) as a post hoc test on all $r, s$ pairs to check for the source of significant difference. Such pairwise comparisons also allow us to find cliques---a clique is a set of algorithms where the pairwise test does not reject between any pair. Cliques need not be disjoint.

Remember that since we are doing post hoc tests on all $r, s$ pairs to find cliques (a total of $L(L-1)/2$ tests), to retain an overall significance level of $\alpha$, the significance level of each post hoc test should be corrected.

\section{Case Studies}
\label{sec-exp}

Using state of the art learning algorithms on several data sets, we do a set of experiments to show the uses of the multivariate tests and their comparison with the univariate tests; we also show how new measures can be extracted by multivariate analysis.

\subsection{Setup}

We use the following seven algorithms:
\begin{itemize}
\item {\tlda}: Linear discriminant classifier. 
\item {\tqda}: Quadratic discriminant classifier. 
\item {\tknn}: $k$-nearest neighbor with $k = 10$.
\item {\tct}: C4.5 decision tree. 
\item {\trf}: Random forest is an ensemble of decision trees.
\item {\tsvmone}: Support vector machine (SVM) with a linear kernel \citep{libsvm}.
\item {\tsvmtwo}: SVM with a quadratic kernel.
\end{itemize}

We use four bioinformatics data sets, namely, {\em acceptors} \citep{kulp96}, {\em donors} \citep{kulp96}, {\em arabidopsis} \citep{pedersen97}, {\em polyadenylation} \citep{liu03}. We also use the {\em ec} data set \citep{qiu07} where data is provided as a set of kernel matrices, instead of a feature-based representation. On all data sets, we use 10-fold cross-validation, set $\alpha=0.05$ in the statistical tests, and use Holm's correction \citep{holm79} for multiple comparisons.

Before starting any comparison, first, we check if the multivariate normality assumption holds and for this, we use the multivariate test of normality \citep{mardia70} to the results of all of our classifiers on all data sets. Indeed we see that the test never rejects the normality of (tpr, fpr) or (precision, recall) results. This shows that the use of multivariate tests that assume normality are applicable in these experiments.

\subsection{Univariate Test on Error vs. Multivariate Test on (tpr, fpr)}

\begin{figure}[ht]
\begin{center}
\begin{tabular}{cc}
(a) & (b) \\
\includegraphics[width=0.5\linewidth]{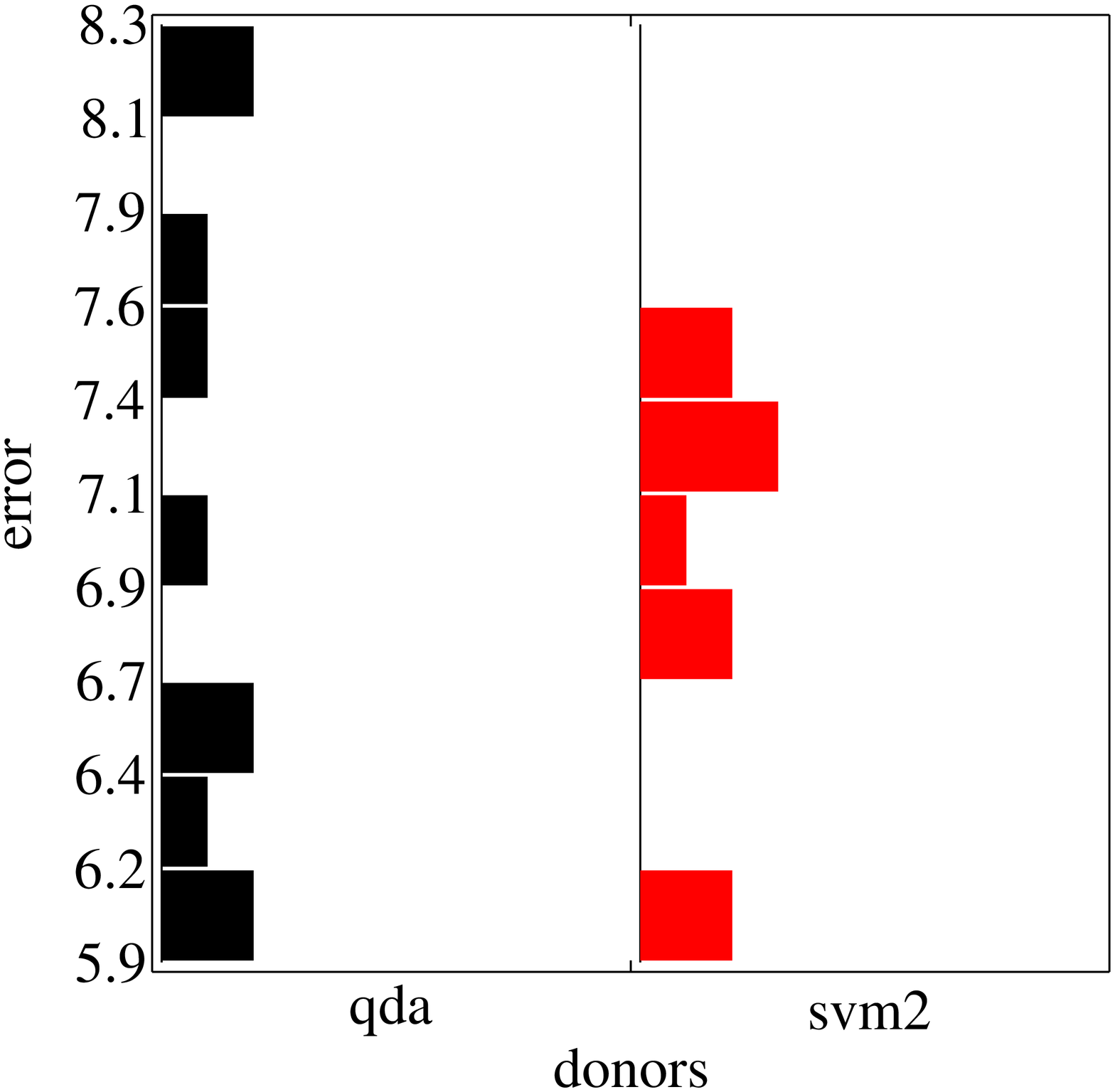} & \includegraphics[width=0.5\linewidth]{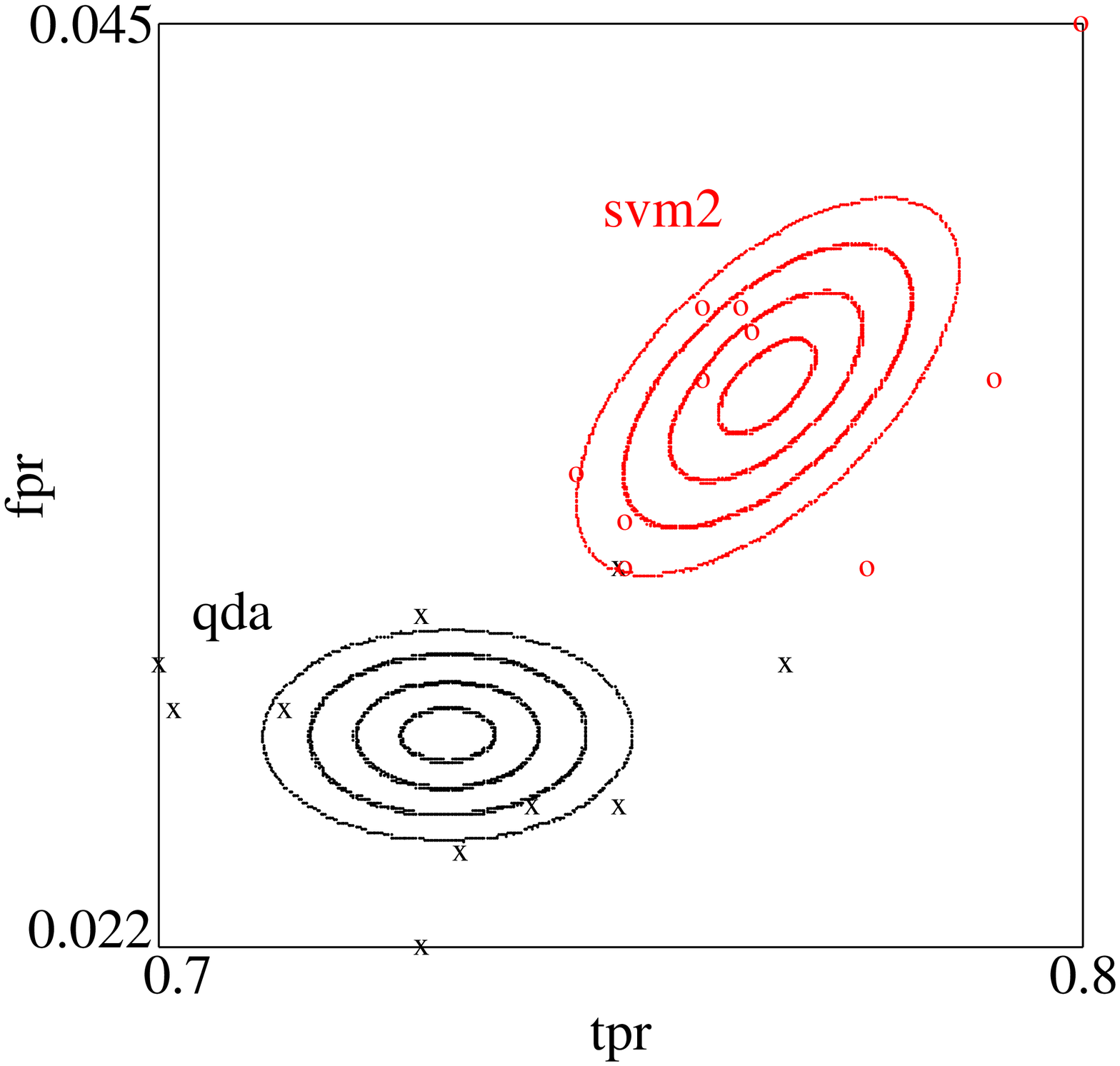}   \\
\end{tabular}
\end{center}
\caption{On the {\em donors} data set when comparing {\tqda} and {\tsvmtwo}, error histograms are given in (a) and (tpr, fpr) scatter values and the isoprobability contours of the fitted 2-variate Gaussians are given in (b). The histograms in (a) overlap and hence the univariate test says they have comparable error, whereas the two dimensional densities in (b) are well-separated and hence the multivariate test rejects.}
\label{fig-univsmulti}
\end{figure}

As our first case study, we compare the univariate $t$ test on misclassification error with our proposed multivariate test on (tpr, fpr). We use {\tqda} and {\tsvmtwo} on the {\em donors} data set. Figure \ref{fig-univsmulti}(a) shows that the two algorithms have comparable error histograms (remember that because we use 10-fold cross-validation, we have ten results for each algorithm) and hence, the univariate test on error does not reject the null hypothesis that the expected errors are equal. 

In Figure \ref{fig-univsmulti}(b), we see the (tpr, fpr) scatter plots of the two methods and the isoprobability (equal probability) contours of the fitted bivariate Gaussians. We observe that {\tsvmtwo} has higher fpr and {\tqda} has lower tpr, that is, higher false negative rate. The two densities are well-separated and that is why Hotelling's multivariate test on (tpr, fpr) rejects the null hypothesis that the expected performances are equal.  

That is, {\tsvmtwo} has more false positives and {\tqda} has more false negatives, but overall, their sum (error) are comparable. The sources of error are different for the two algorithms and the multivariate test can detect this whereas the univariate test on error cannot. So if we only care about the misclassifications, a univariate test on error is sufficient but if we want to make a distinction between the false negatives and false positives---and we generally do---we need to use a multivariate test on the two.

Then we do this comparison on all four data sets and for all pairs of algorithms where we compare the do not reject/reject decisions of the univariate test on error and the multivariate test on (tpr, fpr). We have a total of $4\times 7\times 6/2=84$ pairs and we do each comparison 10 times, running 10-fold cv 10 times independently, hence we have a total of 840 results. The breakdown of percentages are given in Table~\ref{tab-errvstprfpr}. We see that though the two tests agree on a majority of the cases, the multivariate tends to reject more, that is, finds a difference, and we see that in almost 15 percent of the cases, the multivariate test rejects whereas the univariate test on error finds no difference and does not reject. The opposite rarely happens. 

\begin{table}[!h]
\caption{Overall pairwise comparison percentages of the do not reject/reject decisions of the univariate test on error and the multivariate test on (tpr, fpr).}
\begin{center}
\begin{tabular}{lrrr}
\hline
			& \multicolumn{2}{c}{Multivariate} & \\
Univariate 	&   Do not reject  &  Reject  & Total \\ 
\hline
Do not reject   	&   7.80 & 14.75 & 22.55 \\
Reject   	&   1.52 & 75.93 & 77.45 \\
Total 		&   9.32 & 90.68 & 100.00 \\
\hline
\end{tabular}
\end{center}
\label{tab-errvstprfpr} 
\end{table}

\subsection{Univariate Test on $F$ Measure vs. Multivariate Test on (Precision, Recall)}

\begin{figure}[t!]
\begin{center}
\begin{tabular}{cc}
(a) & (b) \\
\includegraphics[width=0.5\linewidth]{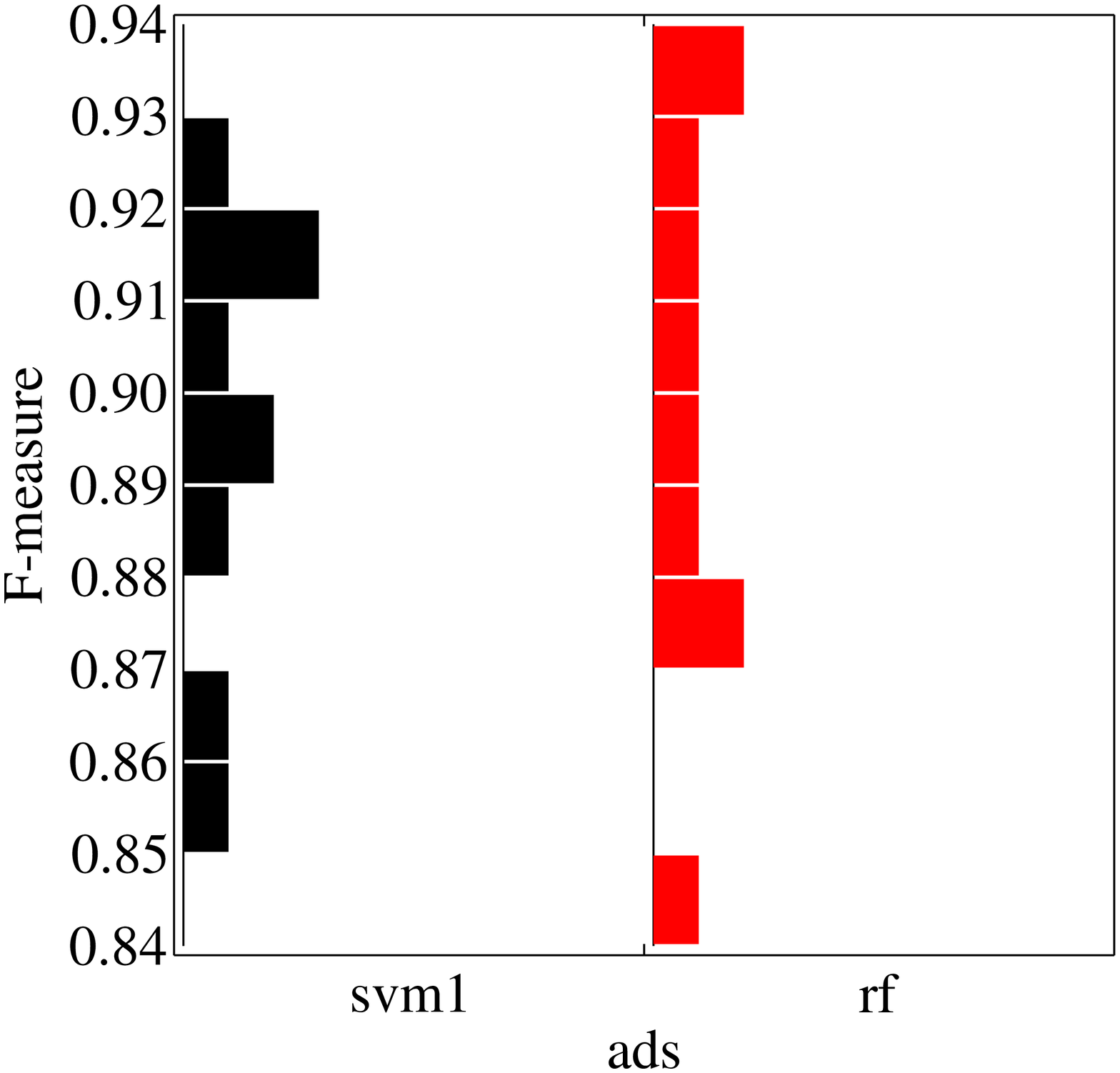} & \includegraphics[width=0.5\linewidth]{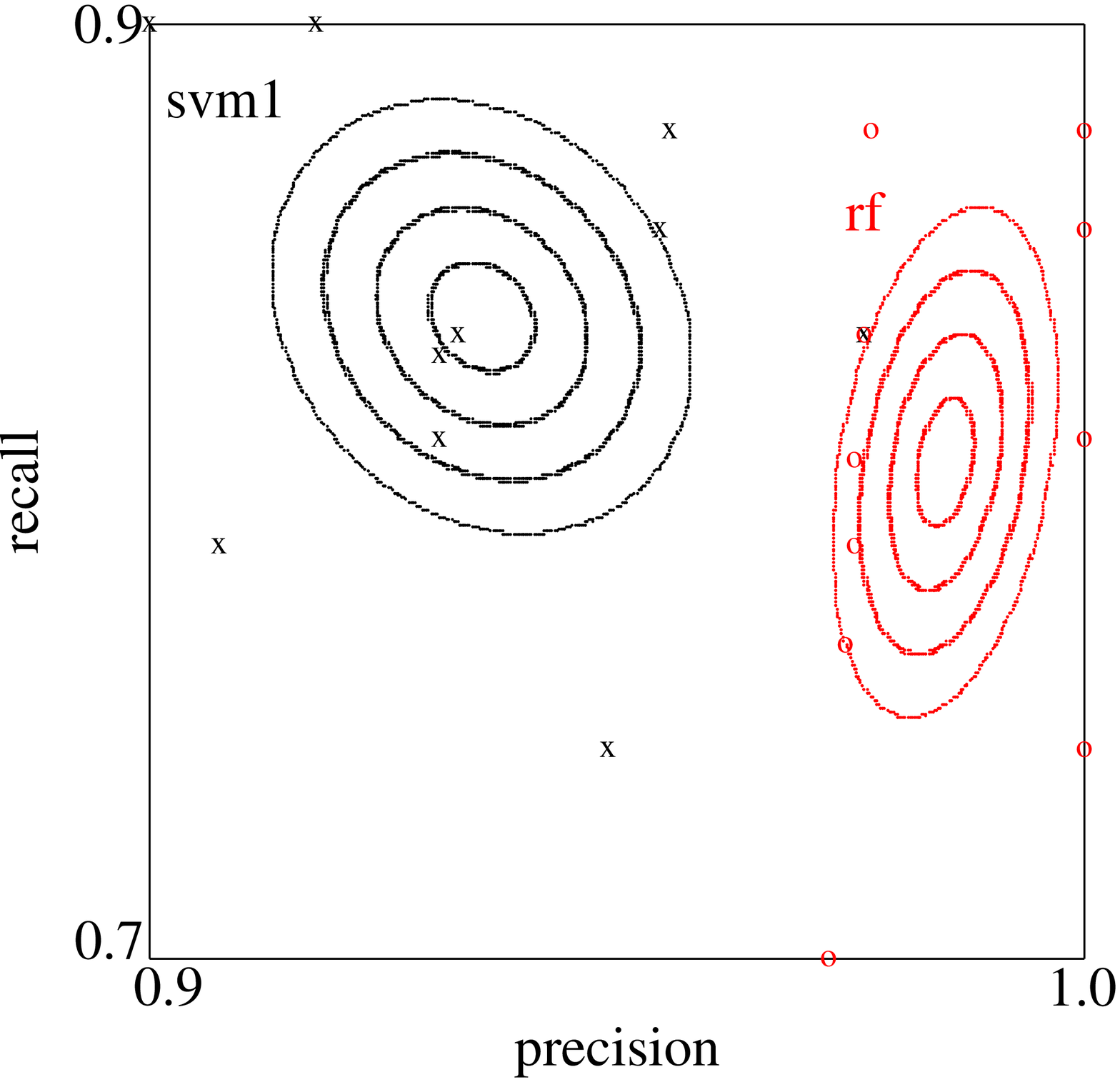}   \\
\end{tabular}
\end{center}
\caption{On the {\em arabidopsis} data set, when we compare {\tsvmone} and {\trf}, the histograms of the univariate $F$ measure values are given in (a), and (precision, recall) scatter plots and the contours of the fitted Gaussians are given in (b). The univariate $F$ measure histograms overlap and the univariate test finds no difference, whereas the two-dimensional Gaussians are well-separated and the Hotelling's multivariate test rejects.}
\label{fig-univsmulti3}
\end{figure}

In information retrieval, we concentrate on the positive class and instead of tpr and fpr, we use precision and recall. $F$ measure combines precision and recall in a single value and is frequently used to assess algorithms. In this case study, we compare the univariate $t$ test on $F$ measure with our proposed multivariate test on (precision, recall). We use {\tsvmone} and {\trf} on the {\em arabidopsis} data set. Figure \ref{fig-univsmulti3}(a) shows that the algorithms have comparable $F$ measure histograms and the univariate test finds no difference. But if we look at Figure \ref{fig-univsmulti3}(b) where we see the (precision, recall) scatter plots and the contour plots of the fitted bivariate Gaussians, we see that they have comparable recall but {\trf} has higher precision, and that is why Hotelling's multivariate test on (precision, recall) rejects the null hypothesis that the means are the same. Again, we see that the multivariate test can detect a difference which is lost to the univariate test.

Then, we compare the do not reject/reject decisions of the univariate test on $F$ measure and the multivariate test on (precision, recall) on all four data sets for all pairs of algorithms. The breakdown of 840 comparisons for this case is given in Table~\ref{tab-Fvsprecrec}. Here too, we see that though the two tests agree on a majority of the cases, the multivariate tends to reject more, that is, finds a difference, and we see that in around 24 percent of the cases, the multivariate test rejects whereas the univariate test on $F$ measure finds no difference and does not reject. The opposite, again, rarely happens. 

\begin{table}[!h]
\caption{Overall pairwise comparison percentages of the do not reject/reject decisions of the univariate test on $F$ measure and the multivariate test on (precision, recall)}
\begin{center}
\begin{tabular}{lrrr}
\hline
			& \multicolumn{2}{c}{Multivariate} & \\
Univariate 	&   Do not reject  &  Reject  & Total \\ 
\hline
Do not reject   	&   7.46 & 24.07 & 31.53 \\
Reject   	&   1.69 & 66.78 & 68.47 \\
Total 		&   9.15 & 90.85 & 100.00 \\
\hline
\end{tabular}
\end{center}
\label{tab-Fvsprecrec} 
\end{table}

\subsection{Comparing Kernels using a Multivariate Test}

A statistical test can be used to compare not only algorithms but also hyper parameters of the same algorithm. In SVM for example, the most important hyper parameter is the kernel and different kernels may lead to different behaviors. Defining kernels is a good way to incorporate prior knowledge in machine learning, and application-specific kernels are defined and used successfully in various bioinformatics applications \citep{scholkopfetal04}; it is hence vital to be able to detect differences in performance due to kernels, and our proposed multivariate test can be used for this.

In this case study, we compare two SVM classifiers on the {\em ec} data set \citep{qiu07} but with different kernels using different representations; one is the {\em vector\/} kernel where each protein is described by a vector of 55 features, and the other is the {\em contact\/} kernel which captures the pairwise and multi-body interactions among amino acids in a protein structure.

\begin{figure}[h!]
\begin{center}
\begin{tabular}{cc}
(a) & (b) \\
\includegraphics[width=0.5\linewidth]{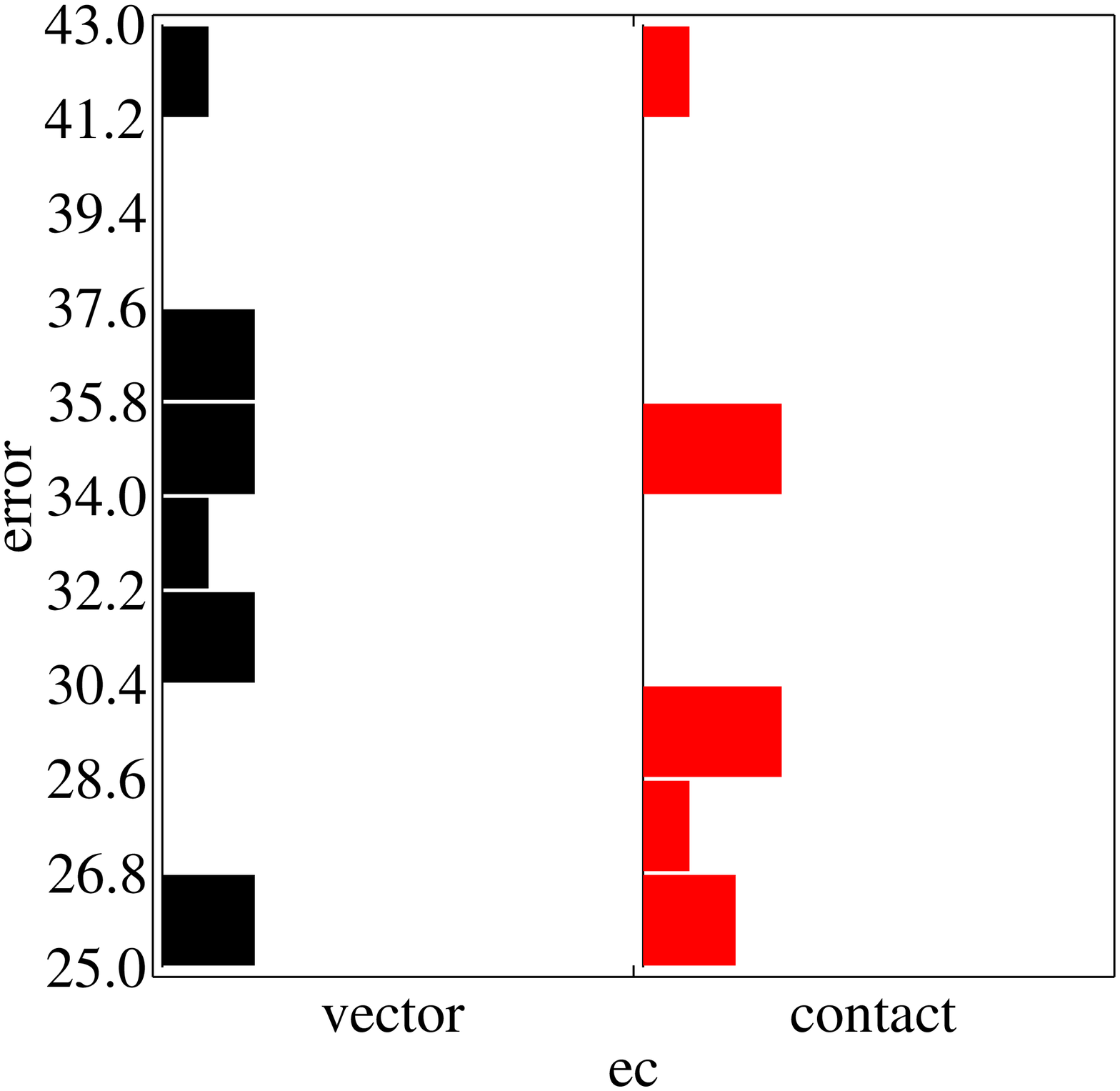} & \includegraphics[width=0.5\linewidth]{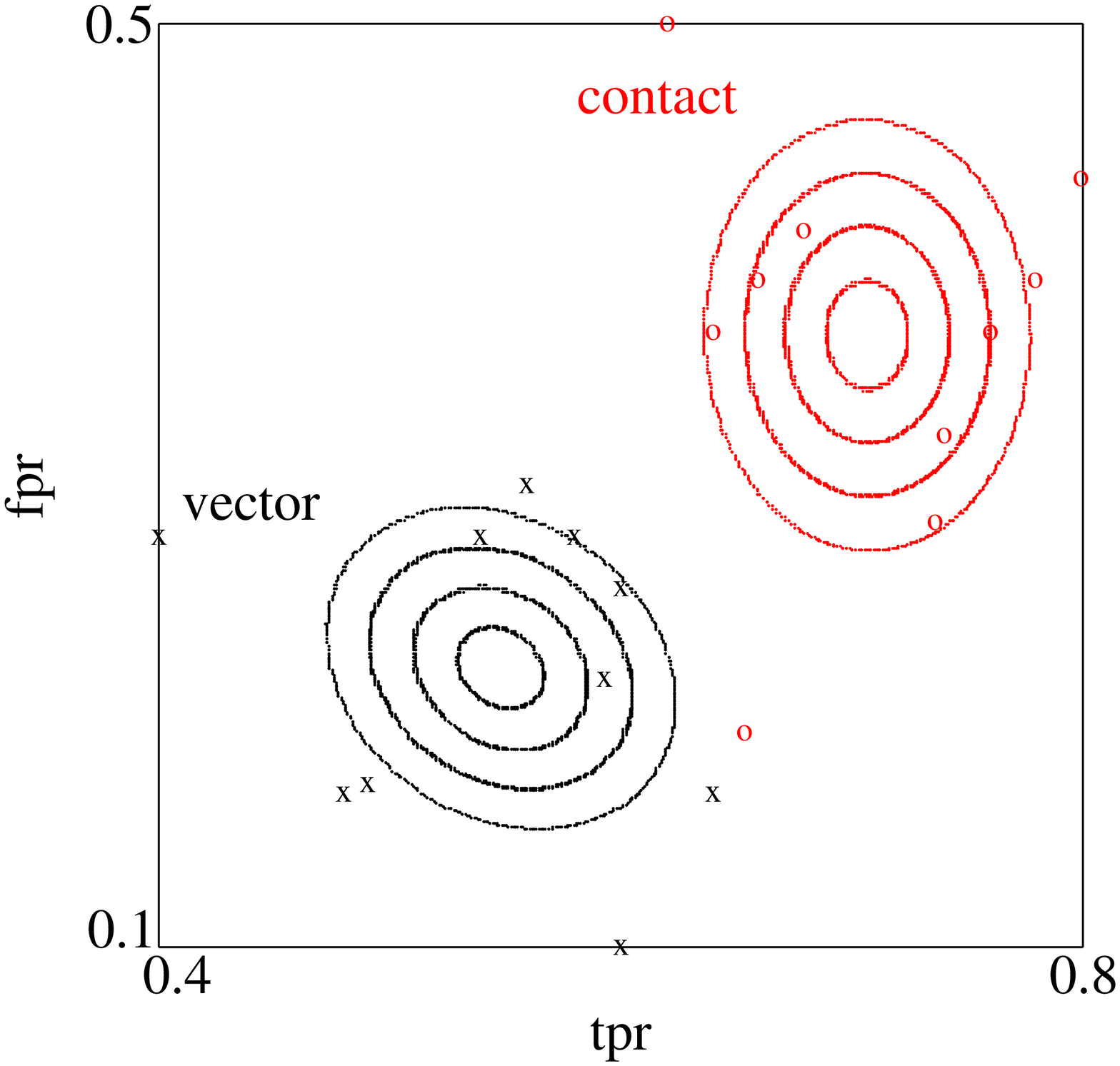}   \\
\end{tabular}
\end{center}
\caption{On {\em ec} data set, we compare two support vector machine classifiers with different kernels, namely {\em vector\/} and {\em compact\/} kernels. In (a), we see that the error histograms overlap considerably and hence the univariate test does not reject, whereas in (b), where we see the scatter plots of (tpr, fpr) and contours of the fitted 2-variate Gaussians, the two distributions are quite apart and the multivariate test rejects.}
\label{fig-univsmulti-kernel}
\end{figure}

In Figure \ref{fig-univsmulti-kernel}(a), we see that the two SVMs with different kernels lead to comparable error histograms and the univariate test does not reject, but if we look at Figure \ref{fig-univsmulti-kernel}(b), we see that SVM with the {\em contact\/} kernel has higher fpr, and SVM with the {\em vector\/} kernel has lower tpr, that is, higher false negative rate. Hotelling's multivariate test on (tpr, fpr) finds a difference between the two kernels that cannot be detected if we simply compared in terms of misclassification error.

\subsection{Finding Multivariate Cliques}

\begin{figure}[h!]
\begin{center}
\begin{tabular}{cc}
(a) & (b) \\
\includegraphics[width=0.5\linewidth]{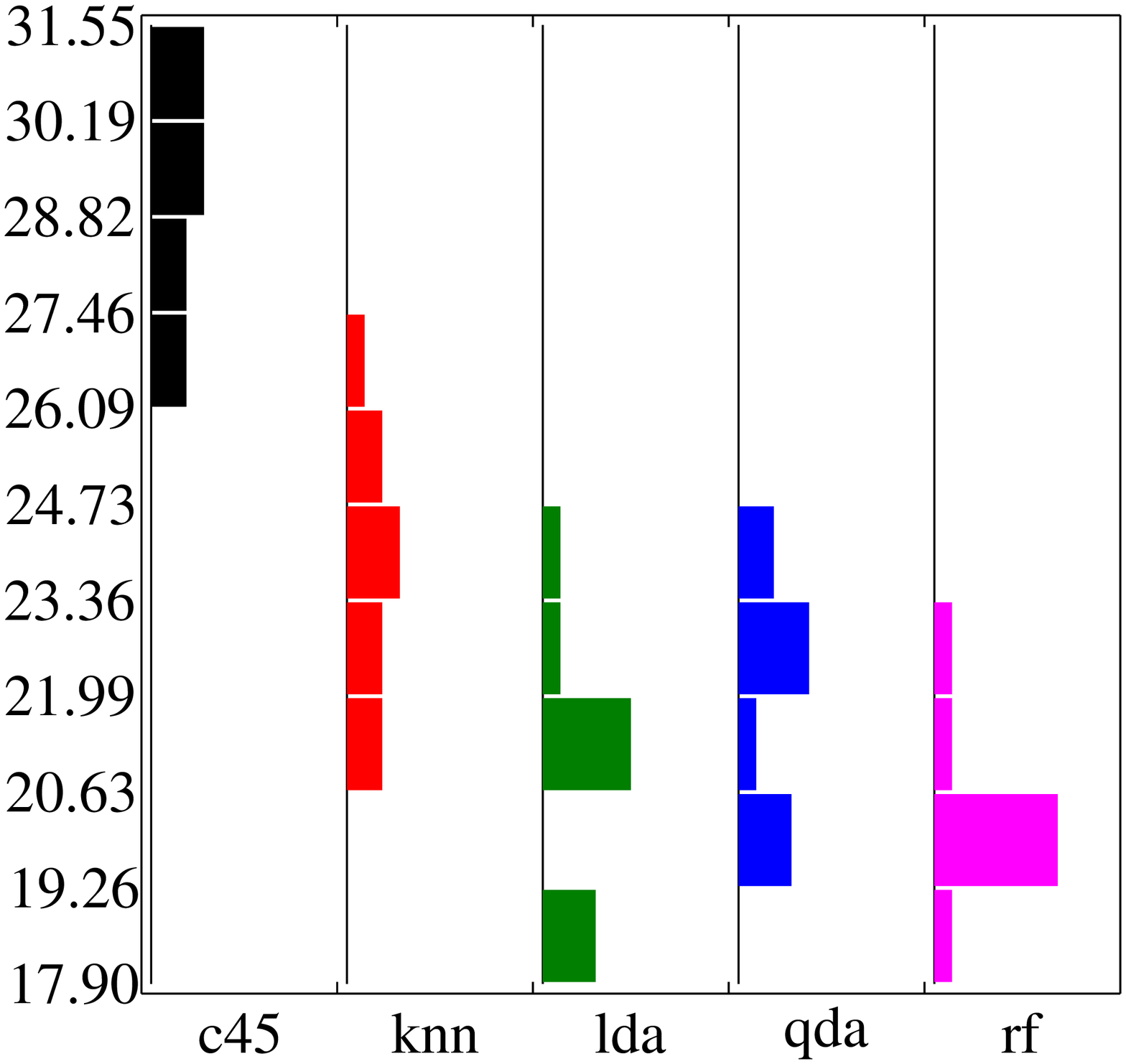} & \includegraphics[width=0.5\linewidth]{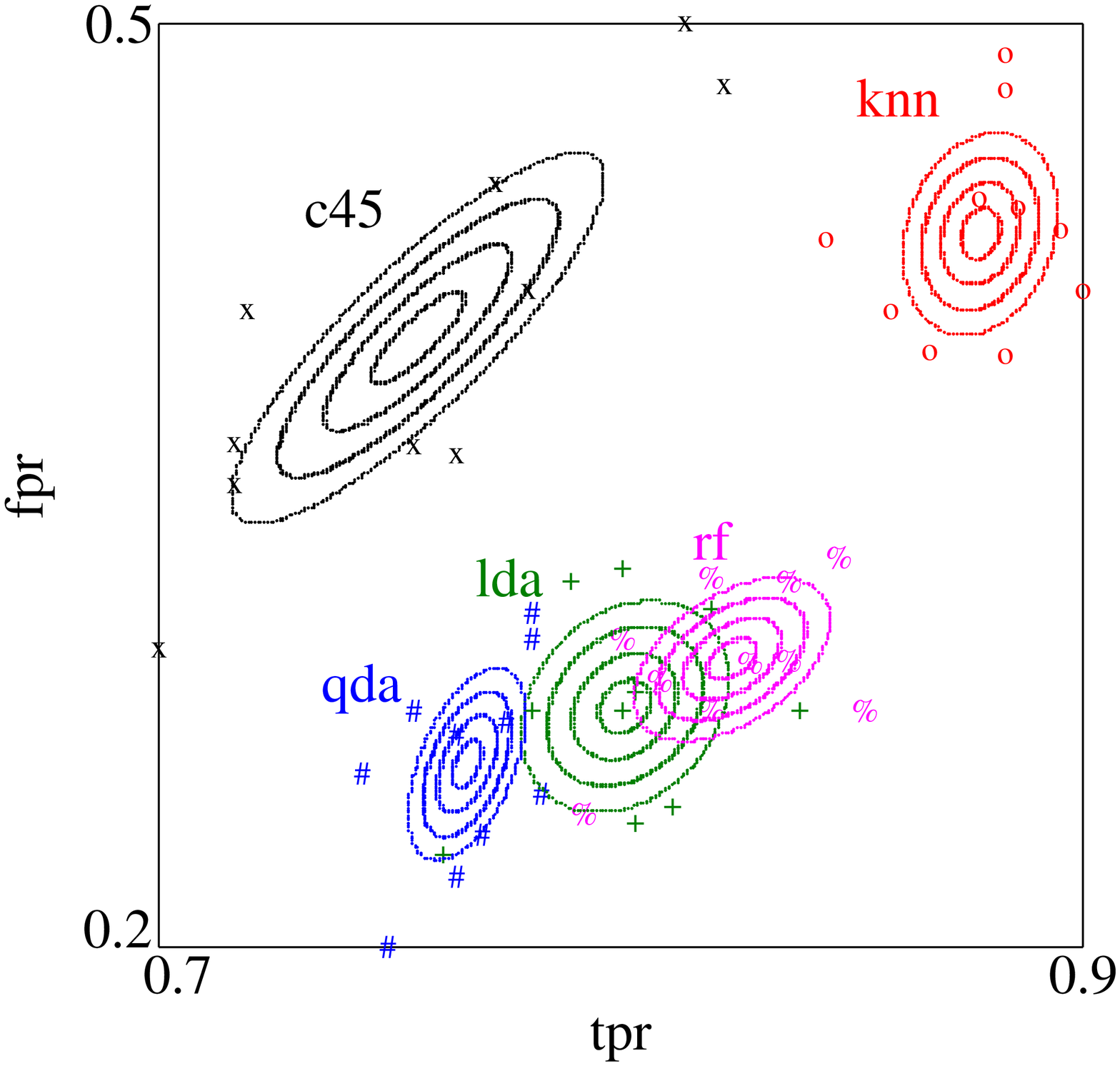}   \\
\end{tabular}
\end{center}
\caption{Comparison of five algorithms on {\em polyadenylation}, where (a) shows the error histograms of the five classifiers, (b) the scatter plots and the contour plots of the fitted bivariate Gaussians with respect to (tpr, fpr).}
\label{fig-multiple}
\end{figure}

Let us now consider a case study where we have $L>2$ algorithms. Again, let us compare error and (tpr, fpr). We run {\tct}, {\tlda}, {\trf}, {\tqda}, and {\tknn} on the {\em polyadenylation} data set. The error histograms are given in Figure \ref{fig-multiple}(a) and multivariate (tpr, for) plots in Figure \ref{fig-multiple}(b).

For the univariate case with error, we first do ANOVA, it rejects and we find an ordering of the algorithms by using post hoc univariate tests:

\def\lab{\hspace{-.38in}\raisebox{-.025in}{\rule{.38in}{.005in}}}
\def\labb{\hspace{-.58in}\raisebox{-.025in}{\rule{.58in}{.005in}}}
\def\lbb{\hspace{-.52in}\raisebox{-.05in}{\rule{.52in}{.005in}}}
\def\lbc{\hspace{-.87in}\raisebox{-.05in}{\rule{.87in}{.005in}}}

{\trf}\ {\tlda}\lab\ {\tqda}\lbb\ {\tknn}\ {\tct}

The algorithms are sorted in terms of average error and an underline under two or more methods denote that there is no significant difference. Here for example {\trf} has the lowest average error and {\tct} has the highest; there is no significant difference between {\trf} and {\tlda}, but there is between {\trf} and {\tqda}.

With multivariate (tpr, fpr) (see Figure \ref{fig-multiple}(b)), first  we use MANOVA and it rejects. We then do pairwise multivariate tests and find cliques as: \{{\tct}, ({\tlda}, {\trf}), {\tqda}, and {\tknn}\}; that is, there is no significant multivariate difference between {\tlda} and {\trf}; otherwise they are all distinct.

We can also do post hoc univariate tests along the dimensions separately. For tpr, the ordering we find is

{\tknn}\ {\trf}\ {\tlda}\lab\ {\tqda}\ {\tct}\lbc

\noindent and the ordering we find with respect to fpr is

{\tqda}\ {\tlda}\lbb\ {\trf}\lab\ {\tct}\ {\tknn}\labb

{\tlda} and {\trf} which make up a clique in (tpr, fpr) are underlined together in both tpr and fpr. Note how the multivariate analysis gives us much more information about the behavior of the algorithms than a comparison that just uses the misclassification error.

\subsection{Extracting New Performance Measures}

Let us see now another type of multivariate analysis we can do once we have multivariate data. Here, the idea is to use directly the four values, {\etp}, {\efn}, {\efp}, and {\etn}, and learn a performance measure from them, rather than using predefined measures such as tpr, precision, $F$ measure, and so on. As we discuss in Section~\ref{sec-L}, given the four-dimensional data from $k$ folds of all $L>2$ algorithms, we calculate the $\mH$ and $\mE$ matrices and find the eigenvectors of $\mE^{-1}\mH$. The projections with these eigenvectors give us the new performance measures learned as linear combinations of the original {\etp}, {\efn}, {\efp}, and {\etn}.

\begin{figure}[h!]
\begin{center}
\begin{tabular}{cc}
(a) & (b) \\
\includegraphics[width=0.5\linewidth]{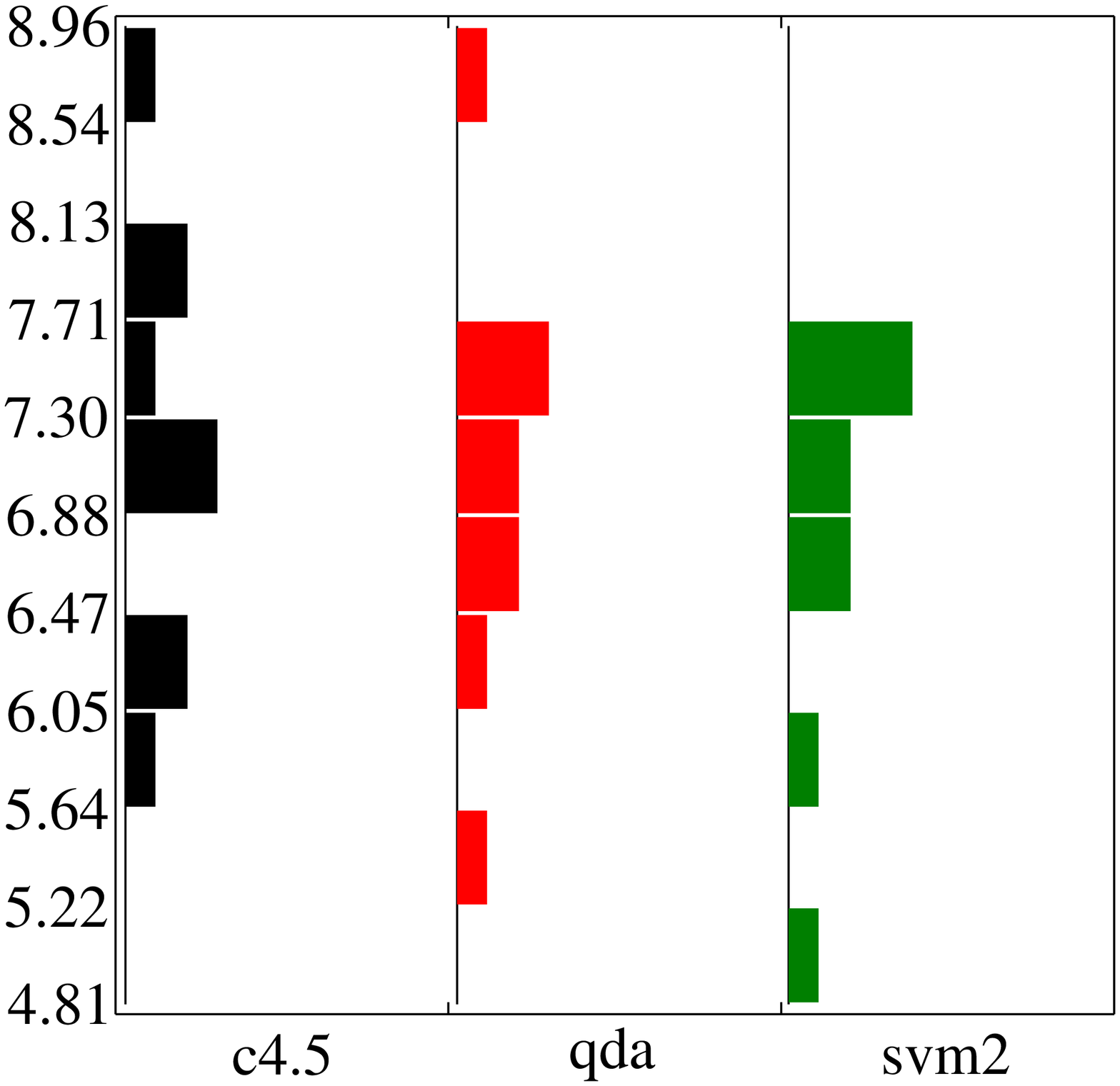} & \includegraphics[width=0.5\linewidth]{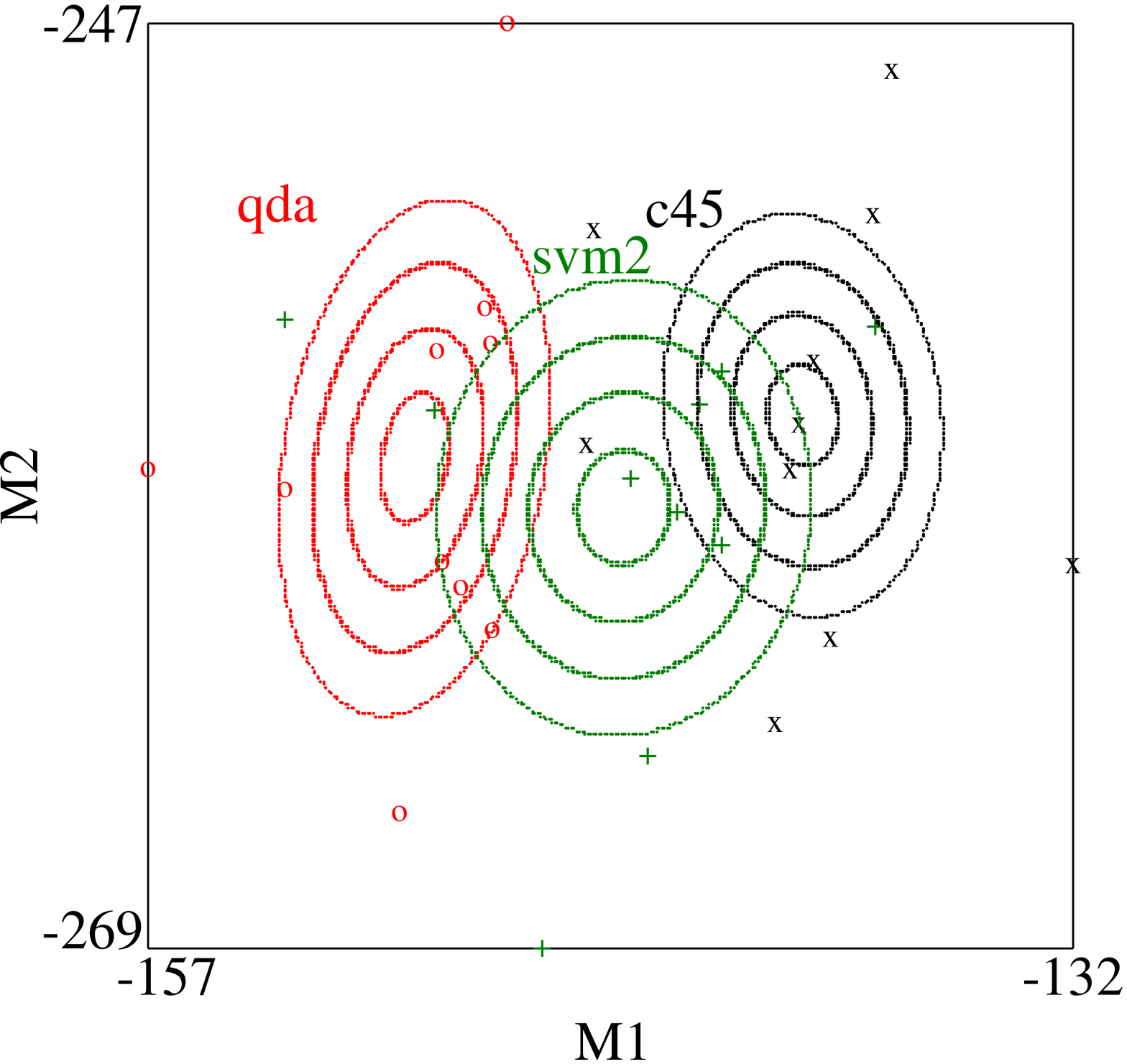} \\ 
(c) & \\
\includegraphics[width=0.5\linewidth]{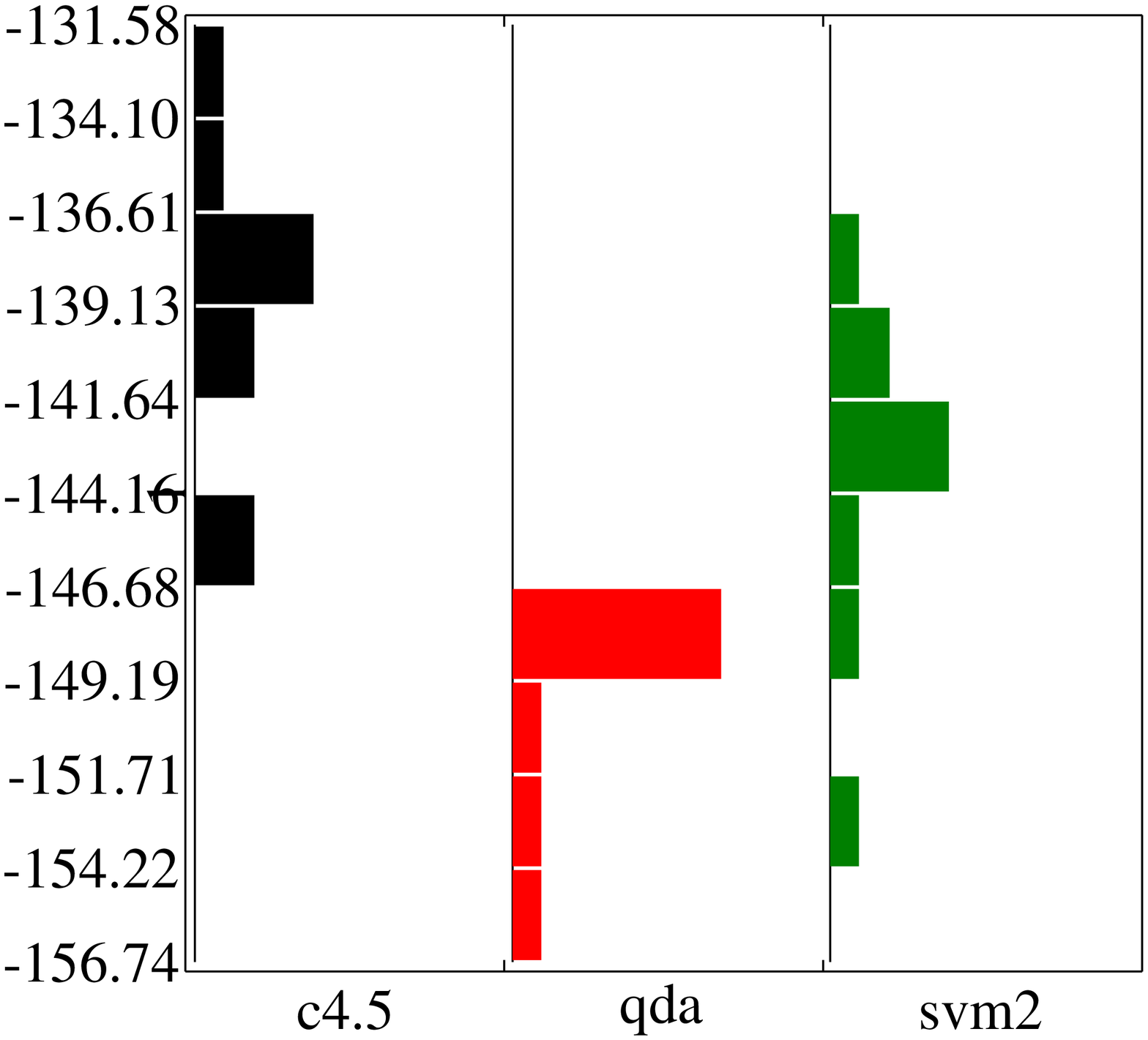} & \\
\end{tabular}
\end{center}
\caption{Comparison of three selected algorithms {\tct}, {\tqda}, and {\tsvmtwo} on {\em donors} dataset. (a) shows the error histograms of the three classifiers, (b) the scatter plots and the contour plots of the fitted bivariate Gaussians with respect to the two directions found by multivariate analysis, and (c) the histogram of the projection on the direction that best separates the three in terms of (\eTP, \eFP, \eTN, \eFN)---this is the M1 axis of (b).}
\label{fig-spect-eig}
\end{figure}

Let us see an example. On {\em donors}, we compare three algorithms, namely, {\tct}, {\tqda}, and {\tsvmtwo}. In Figure~\ref{fig-spect-eig}(a), we see the error histograms of the three classifiers and the univariate test cannot find any difference between them---ANOVA does not reject. 

The new learned measures given by the two eigenvectors of $\mE^{-1}\mH$ are
\begin{eqnarray}
M1	&:& -0.379\cdot tp + 0.733\cdot fp - 0.206\cdot tn - 0.525\cdot fn \nonumber \\
M2	&:& -0.098\cdot tp + 0.323\cdot fp - 0.573\cdot tn + 0.746\cdot fn \nonumber 
\end{eqnarray}

The eigenvalues are $\lambda_1=19.995$ and $\lambda_2=7.731$ respectively. The plot of the three classifier results after projection in this new (M1, M2) space is given in Figure~\ref{fig-spect-eig}(b). 

By looking at the magnitude of the projection weights, we can extract information. For example looking at M1, we can say that the four algorithms differ most in terms of {\eFP} and {\eFN} (that is, in terms of their wrong predictions on the positive and negative instances), and less in terms of {\eTP} or {\eTN} (that is, their true decisions). Similarly, M2 gives higher weight to {\eTN} and {\eFN} (that is, in terms of the assignments to the negative class). Information like this enables a better analysis of the results and hence a better assessment and comparison of classification algorithms.

We can do 2-variate test to compare the three methods in the (M1, M2) space, but knowing that M1 explains 73 per cent of the variance ($\lambda_1/(\lambda_1+\lambda_2)$), it suffices. In Figure~\ref{fig-spect-eig}(c), we see the histogram of the three algorithms on M1 only. As we can see, the three algorithms are well-separated and indeed the univariate test according to M1 finds a significant difference between all pairs. 

So to sum up, what we are doing here is that instead of assuming any predefined measure, we are using multivariate analysis to learn the best measure from the raw confusion-matrix data and then use a test according to these learned measure.

In this case with three algorithms, a single eigenvector may be sufficient; in case where there are many algorithms, we can use multiple eigenvectors and then do a multivariate test in this new space of projections. Another advantage is that the eigenvectors are orthogonal and hence the learned performance measures will be independent.

\section{Conclusions}
\label{sec-conc}

We use multivariate tests to compare the performances of classification algorithms. We consider multiple performance measures simultaneously, without needing to sum them up in a cumulative measure such as misclassification error or $F$ measure, which may hide differences in the behavior of the algorithms. 

We proposed the use of multivariate tests in our previous work \citep{lion5} where we gave the motivation for multivariate comparison and compared error with (tpr, fpr) on datasets from the UCI repository. In this present paper, we extend the use of multivariate comparison by (i) comparing univariate $F$-measure with multivariate (precision, recall), (ii) comparing kernels in support vector machines---that is, not only for comparison of algorithms but also hyper parameters, (iii) the use of multivariate analysis to learn the best discriminating measure from data rather than pre-defined measures, and (iv) experiments on real-world bioinformatics data sets.

In the literature, various ways have been proposed to combine multiple measures in a single number. \cite{caruana04} compare different performance metrics such as accuracy, lift, $F$ score, area under the ROC curve, average precision, precision/recall break-even point, squared error, cross entropy, and probability calibration; they show that these metrics are correlated and propose a new measure that they name SAR as the average of Squared error, Accuracy and area under the Roc curve. \cite{bhowan12} combine true positive rate and true negative rate as a new performance metric and use their weighted average in a fitness function for classification with unbalanced data. \cite{seliya09} combine them taking their correlation into account. Note that all of these at the end give a single measure which hides the differences in the sources; what we propose is to test using multiple measures without needing to compress them in a single value. 

We discuss and show in case studies the use of multivariate tests for comparing (tpr, fpr) and (precision, recall). Another pair of measures is sensitivity and specificity, frequently used in medical applications, and though we have not in this work, multivariate tests can also be used with (sensitivity, specificity). 

The tests we use can compare an arbitrary number of algorithms and allow finding multivariate cliques, that is subsets of algorithms among which there is no significant multivariate difference. A multivariate analysis also allows us to construct linear combinations of basic performance statistics that reveals the best way they should be combined to maximize the difference between the behavior of the algorithms.  

We point out that the multivariate test we propose (as well as the univariate tests that are already being used) are not restricted to use after $k$-fold cv and they can follow any improved resampling scheme, should such a scheme be proposed. For example, \cite{dietterich98} has found $5\times 2$ cv to be better than 10-fold cv for comparing according to error and it will be interesting to do a similar comparison for the multivariate case.

In our experiments, we have never witnessed such a case, but sometimes when there are few folds or smaller and noisier data sets, results may not be normally distributed. In such a case, the use of rank data and nonparametric testing is recommended \citep{montgomery09}. It is known that parametric tests are to be preferred over their nonparametric counterparts if their assumptions are satisfied, because they define tighter intervals. But when we are concerned with the normality of our data or the effect of outliers, the use of a nonparametric test for comparison would be an interesting future research direction.

\section*{Acknowledgments}

This work is supported by the Scientific and Technological Research Council of Turkey (T{\"U}B{\.I}TAK) under Grant EEEAG 109E186.

\end{document}